\documentclass[10pt,twocolumn,letterpaper]{article}

\usepackage{iccv}
\usepackage{times}
\usepackage{epsfig}
\usepackage{graphicx}
\usepackage{amsmath}
\usepackage{amssymb}

\usepackage{subfigure}
\usepackage{color}
\usepackage{caption}
\usepackage{array}
\usepackage[normalem]{ulem}

\definecolor{idtColor}{RGB}{0, 114, 189}
\definecolor{cnnColor}{RGB}{153, 122, 0}
\definecolor{originalColor}{RGB}{255, 80, 80}
\definecolor{rbColor}{RGB}{0, 128, 255}
\definecolor{rbActColor}{RGB}{252, 189, 38}
\definecolor{augColor}{RGB}{51, 153, 102}

\newcommand{\Dima}{\textcolor[rgb]{0,0,0}}

\newcommand{\Davide}{\textcolor[rgb]{0,0,0}}

\newcolumntype{C}[1]{>{\centering\let\newline\\\arraybackslash\vspace{3pt}}m{#1}}

\usepackage[breaklinks=true,bookmarks=false]{hyperref}

\iccvfinalcopy % *** Uncomment this line for the final submission

\def\iccvPaperID{1085} % *** Enter the ICCV Paper ID here

\ificcvfinal\fi
\begin{document}

%%%%%%%%% TITLE 
\title{Trespassing the Boundaries:\\ Labeling Temporal Bounds for Object Interactions in Egocentric Video}

\author{Davide Moltisanti \qquad Michael Wray \qquad Walterio Mayol-Cuevas \qquad Dima Damen\\
	University of Bristol, United Kingdom 
	{\tt\tt $\quad$ \small <FirstName>.<LastName>@bristol.ac.uk}}

\maketitle

\pagenumbering{gobble} % this is to remove all page numbers

%%%%%%%%% ABSTRACT
\begin{abstract}
	Manual annotations of temporal bounds for object interactions (i.e. start and end times) are typical training input to recognition, localization and detection algorithms.
	For three publicly available egocentric datasets, we uncover inconsistencies in ground truth temporal bounds within and across annotators and datasets.
	We systematically assess the robustness of state-of-the-art approaches to changes in labeled temporal bounds, for object interaction recognition.
	As boundaries are trespassed, a drop of up to 10\% is observed for both Improved Dense Trajectories and Two-Stream Convolutional Neural Network.
	
	We demonstrate that such disagreement stems from a limited understanding of the distinct phases of an action, and propose annotating based on the Rubicon Boundaries, inspired by a similarly named cognitive model, for consistent temporal bounds of object interactions. Evaluated on a public dataset, we report a 4\% increase in overall accuracy, and an increase in accuracy for 55\% of classes when Rubicon Boundaries are used for temporal annotations.
\end{abstract}

%%%%%%%%% BODY TEXT
\vspace*{-10pt}
\section{Introduction}
\label{sec:introduction}
\vspace*{-3pt}

Egocentric videos, also referred to as first-person videos, have been frequently advocated to provide a unique perspective into object interactions~\cite{kanade2012first,fathi2011understanding, mayol2004interaction}.
These capture the viewpoint of an object close to that perceived by the user during interactions. 
Consider, for example, `turning a door handle'.
Similar appearance and motion information will be captured from an egocentric perspective as multiple people turn a variety of door handles.

Several datasets have been availed to the research community focusing on object interactions from head-mounted~\cite{de2008guide, fathi2011learning, Fathi2012,Damen2014a, lee2012discovering} and chest-mounted~\cite{pirsiavash2012detecting} cameras.
These incorporate ground truth labels that mark the \textit{start} and the \textit{end} of each object interaction, such as `open fridge', `cut tomato' and `push door'.
These temporal bounds are the base for automating action detection, localization and recognition.
They are thus highly influential in the ability of an algorithm to distinguish one {interaction} from another.

As temporal bounds vary, the segments may contain different portions of the untrimmed video from which the action is extracted. 
Humans can still recognize an action even when the video snippet varies or contains only part of the action. Machines are not yet as robust, given that current algorithms strongly rely on the data and the labels we feed to them.
Should these bounds be incorrectly or inconsistently annotated, the ability to learn as well as assess models for action recognition would be adversely affected.

In this paper, we uncover inconsistencies in defining temporal bounds for object interactions within and across three egocentric datasets.
We show that temporal bounds are often ill-defined, with limited insight into how they have been annotated.
We systematically show that perturbations of temporal bounds influence the accuracy of action recognition, for both hand-crafted features and fine-tuned  classifiers, even when the tested video segment significantly overlaps with the ground truth segment.

While this paper focuses on unearthing inconsistencies in temporal bounds, and assessing \Dima{their} effect on object interaction recognition, we take a step further into proposing an approach for consistently labeling temporal bounds inspired by studies in the human mindset.

\vspace*{2pt}
\noindent\textbf{Main Contributions} \hspace{4pt}
More specifically, we:
\begin{itemize}
	\vspace*{-4pt}
	\item Inspect the consistency of temporal bounds for object interactions \textit{across and within} three datasets for egocentric object interactions. We demonstrate that current approaches are highly subjective, with visible variability in temporal bounds when annotating instances of the same action;
	\vspace*{-6pt}
	\item Evaluate the robustness of two state-of-the-art action recognition approaches, namely Improved Dense Trajectories~\cite{Wang2013} and Convolutional Two-Stream Network Fusion~\cite{feichtenhofer2016convolutional}, to changes in temporal bounds. We demonstrate that the recognition rate drops by 2-10\% when temporal bounds are modified albeit within an Intersection-over-Union of more than 0.5;
	\vspace*{-6pt}
	\item Propose, inspired by studies in Psychology, the Rubicon Boundaries to assist in consistent temporal boundary annotations for object interactions;
	\vspace*{-6pt}
	\item Re-annotate one dataset using the Rubicon Boundaries, and show more than 4\% increase in recognition accuracy, with improved per-class accuracies for most classes in the dataset.
\end{itemize}
\vspace*{-6pt}

We next review related works in Section~\ref{sec:relatedWork}, before embarking on inspecting labeling consistenc{ies} in Section~\ref{sec:temporalBoundaries}, evaluating recognition robustness in Section~\ref{sec:experiments} and proposing and evaluating the Rubicon Boundaries in Section~\ref{sec:rubiconBoundaries}. The paper concludes with an insight into future directions. 

\section{Related Work}
\label{sec:relatedWork}
In this section, we review all papers that, up to our knowledge, ventured into the consistency and robustness of temporal bounds for action recognition.

\vspace*{6pt}
\noindent \textbf{Temporal Bounds in Non-Egocentric Datasets} \hspace{4pt}
The leading work of Satkin and Hebert~\cite{satkin2010modeling} first pointed out that determining the temporal extent of an action is often subjective, and that action recognition results vary depending on the bounds used for training. They proposed to find the most discriminative portion of each segment for the task of action recognition. Given a loosely trimmed training segment, they exhaustively search for the cropping that leads to the highest classification accuracy, using hand-crafted features such as HOG, HOF \cite{laptev2008learning} and Trajectons \cite{matikainen2009trajectons}.
Optimizing bounds to maximize discrimination between class labels has also been attempted by Duchenne~\textit{et al.}~\cite{duchenne2009automatic}, where they refined loosely labeled temporal bounds of actions, estimated from film scripts, to increase accuracy across action classes.
Similarly, two works evaluated the optimal segment length for action recognition~\cite{schindler2008action,yang2014effective}.
From the \textit{start} of the segment, {1}-7 frames were deemed sufficient in~\cite{schindler2008action}, with rapidly diminishing returns as more frames were added.
More recently,~\cite{yang2014effective} showed that 15-20 frames were enough to recognize human actions from 3D skeleton joints. 

Interestingly, assessing the effect of temporal bounds is still an active research topic within novel deep architectures.
Recently, Peng~\textit{et al.}~\cite{peng2016multi} assessed how frame-level classifications using multi-region two-stream CNN are pooled to achieve video-level recognition results.
The authors reported that stacking more than 5 frames worsened the action detection and recognition results for the tested datasets, though only compared to a 10-frame stack.

The problem of finding optimal temporal bounds is much akin to that of action localization in untrimmed videos~\cite{wang2016temporal, lea2016segmental, huang2016connectionist}.
Typical approaches attempt to find similar temporal bounds to those used in training, making them equally dependent on manual labels and thus sensitive to inconsistencies in the ground truth labels.

An interesting approach that addressed reliance on training temporal bounds for action recognition and localization is that of Gaidon \textit{et al.} \cite{gaidon2013temporal}. 
They noted that action recognition methods rely on temporal bounds in test videos to be strictly containing an action, and in \textit{exactly} the same fashion as the training segments.
They thus redefined an action as a sequence of key atomic frames, referred to as actoms.
The authors learned the optimal sequence of actoms per action class with promising results. 
More recently, Wang \textit{et~al.}~\cite{Wang_2016_CVPR} represented actions as a transformation from a precondition state to an effect state. The authors attempted to learn such transformations as well as locate the end of the precondition and the start of the effect.
However, both approaches~\cite{gaidon2013temporal, Wang_2016_CVPR} rely on manual \Dima{annotations} of actoms~\cite{gaidon2013temporal} or action segments~\cite{Wang_2016_CVPR}, which are potentially as subjective as the temporal bounds of the {actions themselves}.

\vspace*{6pt}
\noindent \textbf{Temporal Bounds in Egocentric Datasets} \hspace{4pt} Compared to {third} person action recognition (e.g. 101 action classes in~\cite{soomro2012ucf101} and 157 action classes in~\cite{sigurdsson2016hollywood}), egocentric datasets have a smaller number of classes (5-44 classes~\cite{de2008guide, fathi2011learning, Fathi2012, Damen2014a, lee2012discovering, pirsiavash2012detecting, zhou2015temporal}), with considerable ambiguities (e.g. `turn on' vs `turn off' tap). 
Comparative recognition results {have been} reported on these datasets in~\cite{spriggs2009temporal, taralova2011source, Singh16, li2015delving, Ryoo_2015_CVPR, ma2016going}.

Previously, three works noted the challenge and difficulty in defining temporal bounds for egocentric videos~\cite{spriggs2009temporal,Damen2014a,zhou2015temporal}.
In~\cite{spriggs2009temporal}, Spriggs \textit{et al.} discussed the level of granularity in action labels (e.g.~`break egg' vs `beat egg in a bowl') for the CMU dataset~\cite{de2008guide}.
They also noted the presence of temporally overlapping object interactions (e.g. `pour' while `stirring').
In~\cite{wray2016sembed}, multiple annotators were asked to provide temporal bounds for the same object interaction.
The authors showed variability in annotations, yet did not detail what instructions were given to annotators when labeling these temporal bounds.
In~\cite{zhou2015temporal}, the human ability to order pairwise egocentric segments was evaluated as the snippet length varied.
The work showed that human perception improves as the size of the segment increases to~60 frames, then levels off.

\begin{figure*}[t]
	\centering
	\includegraphics[width=\textwidth]{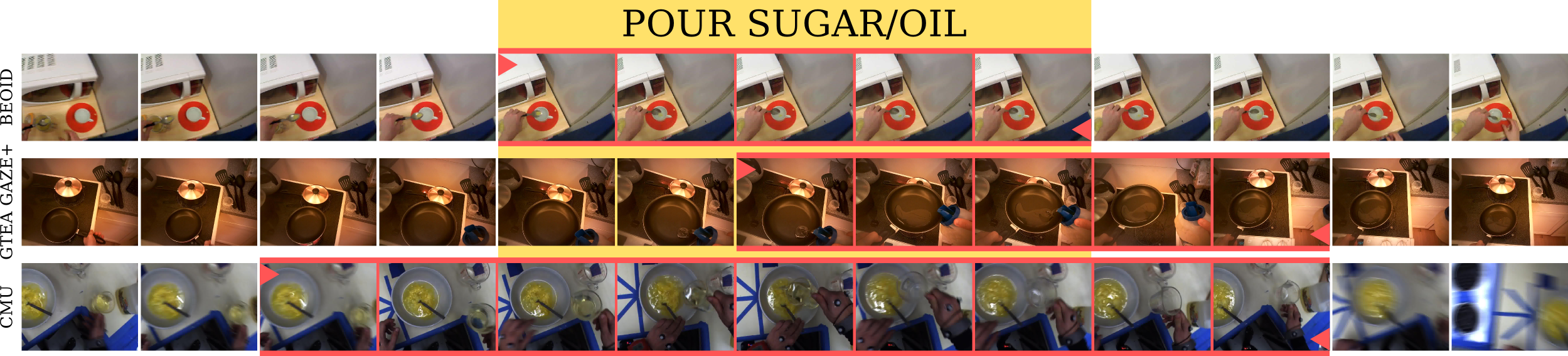}
	\caption{Annotations for action `pour sugar/oil' from BEOID, GTEA Gaze+ and CMU. \Dima{Aligned key frames} are shown along with ground truth annotations (red). The yellow rectangle encloses the motion strictly involved in `pour'.}
	\label{fig:differentBoundariesExample}
\end{figure*}

To summarize, temporal bounds for object interactions in egocentric video have been overlooked and no previous work attempted to analyze the influence of consistency of temporal bounds or the robustness of representations to variability in these bounds.
This paper particularly attempts to answer both questions; \textit{how consistent are current temporal bound labels in egocentric datasets?} and \textit{how sensitive are action recognition results to changes in these temporal bounds?}
We next delve into answering these questions.

\vspace*{-6pt}
\section{Temporal Bounds: Inspecting Inconsistency}
\label{sec:temporalBoundaries}
\vspace*{-2pt}
Current egocentric datasets are annotated for a number of action classes, described using a verb-noun label.
Each class instance is annotated with its label as well as the temporal bounds (i.e.~\textit{start} and \textit{end} times) that delimit the frames used to learn the action model.
Little information is typically provided on how these manually labeled temporal bounds are acquired.
In Section~\ref{subsec:labellingInCurrentDatasets}, we compare labels across and within egocentric datasets.
We then discuss in Section~\ref{subsec:multiple} how variability is further increased when multiple annotators for the same action are employed.

\subsection{Labeling in Current Egocentric Datasets}
\label{subsec:labellingInCurrentDatasets}
We study ground truth annotations for three public datasets, namely BEOID~\cite{Damen2014a}, GTEA Gaze+~\cite{Fathi2012} and CMU~\cite{de2008guide}.
Observably, many annotations base the \textit{start} and \textit{end} of an action as respectively the first and last frames when the hands are visible in the field of view.
Other annotations tend to segment an action more strictly, including only the most relevant physical object interaction within the bounds. Figure~\ref{fig:differentBoundariesExample} illustrates an example of three different temporal bounds for the `pour' action across the three datasets. 
Frames marked in red are those that have been labeled 
%by the different datasets' annotators
{in the different datasets} 
as containing the `pour' action. 
The annotated temporal bounds in this example vary remarkably{;} %among the datasets. 
(i) BEOID's are the tightest; (ii) The start of GTEA Gaze+'s segment is belated: in fact, the first frame in the annotated segment shows {some oil already} in the pan; (iii)
CMU's segment includes picking the oil container {and} putting it down before and after pouring.
These conclusions extend to other actions in the three datasets.

\begin{figure}[t]
	\centering
	\includegraphics[width=1.0\columnwidth]{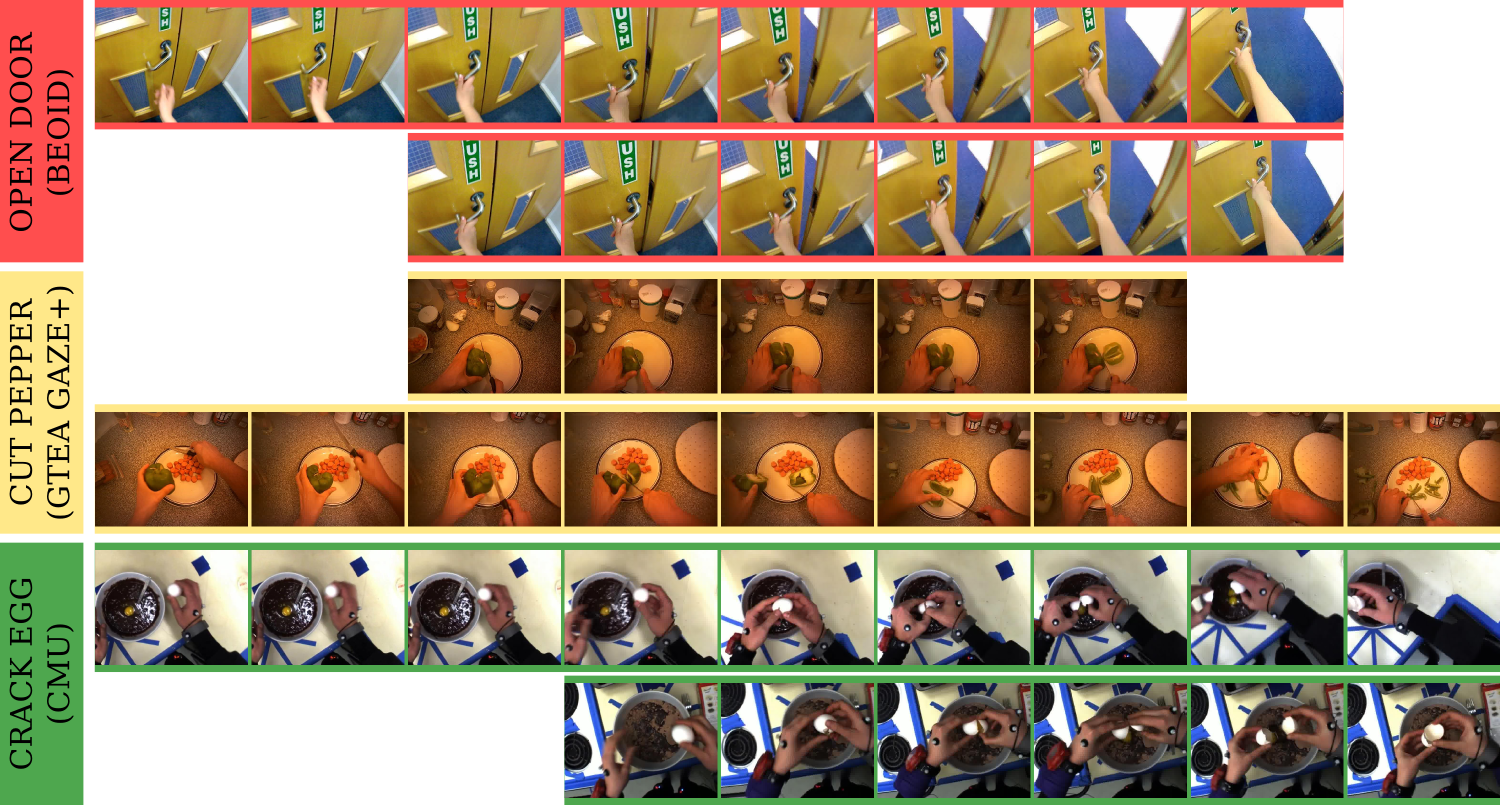}
	\caption{Inconsistency of temporal bounds within datasets. Two segments from each action are shown with considerable differences in start and end times.}
	\label{fig:boundariesInconsistency}
\end{figure}

We observe that annotations are also inconsistent within the same dataset.
Figure \ref{fig:boundariesInconsistency} shows three intra-dataset annotations. (i) For the action `open door' in BEOID, one segment includes the hand reaching the door, while the other starts with the hand already holding the door's handle; (ii) For the action `cut pepper' in GTEA Gaze+, in one segment the user already holds the knife and cuts a single slice of the vegetable. The second segment includes the action of picking up the knife, and shows the subject slicing the whole pepper through several cuts. Note that the length difference between the two segments is considerable - the segments are respectively 3 and 80 seconds long; (iii) For the action `crack egg' in CMU, only the first segment shows the user tapping the egg against the bowl. 

While the figure shows three examples, such inconsistencies have been discovered throughout the three datasets.
However, we generally observe that GTEA Gaze+ shows more inconsistencies, which could be due to the dataset size, as it is the largest among the evaluated datasets.

\vspace*{-4pt}
\subsection{Multi-Annotator Labeling}
\label{subsec:multiple}
\begin{figure}[t]
	\centering
	\includegraphics[width=1\columnwidth]{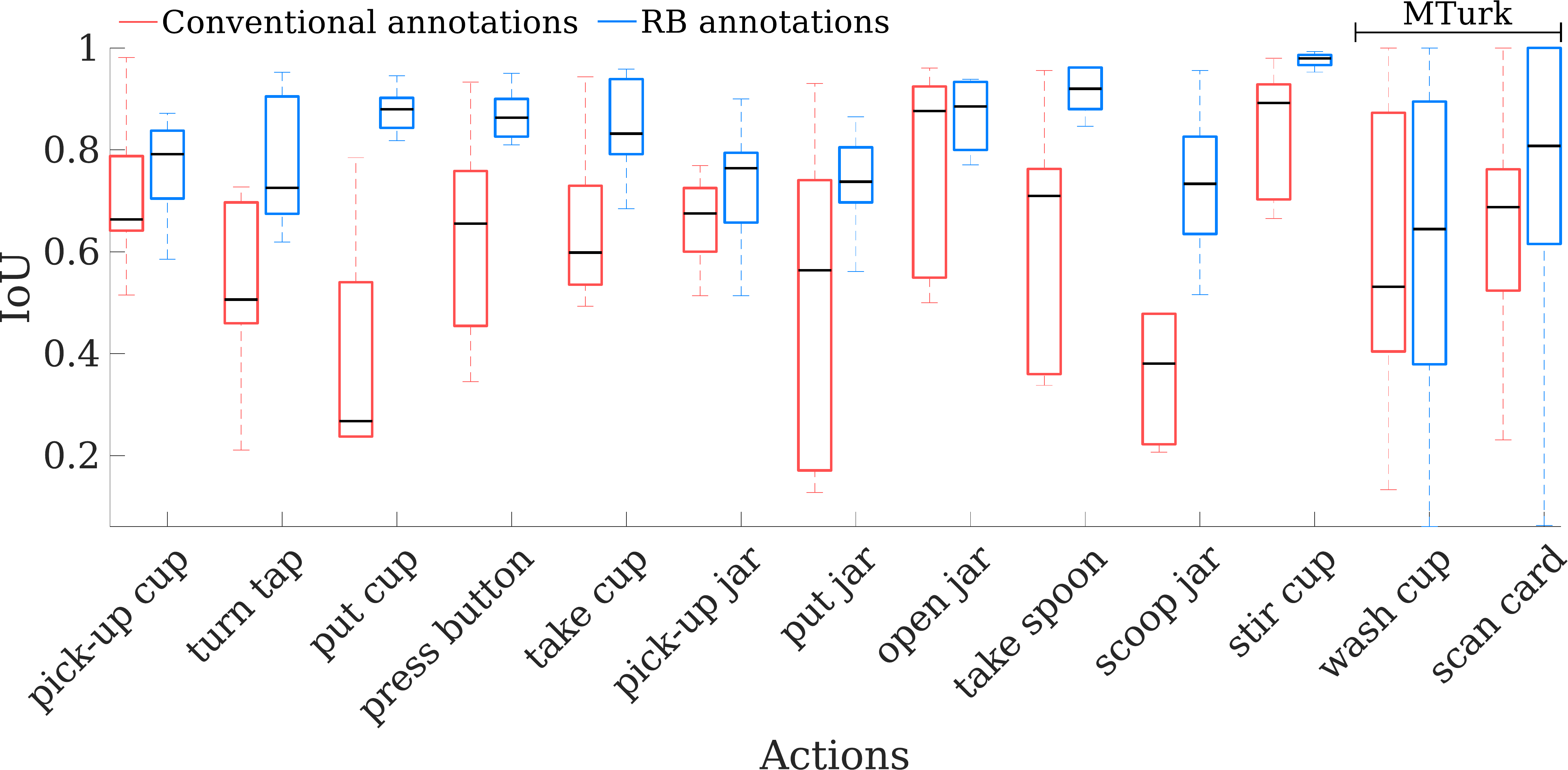}
	\caption{IoU comparison between conventional (red) and RB (blue) annotations for several object interactions.} 
	\label{fig:beoidAnnotations_boxPlots}
\end{figure}

\vspace*{-2pt}
As noted above, defining when an object interaction begins and finishes is highly subjective. 
There is usually little agreement when different annotators segment the same object interaction. 
To assess this variability, we collected 5 annotations for several object interactions from an untrimmed video of the BEOID dataset. First, annotators were only informed of the class name and asked to label the start and the end of the action. We refer to these annotations as \textit{conventional}. We then asked a different set of annotators to annotate the same object interactions following our proposed Rubicon Boundaries (RB) approach which we will present in Section \ref{sec:rubiconBoundaries}.
Figure~\ref{fig:beoidAnnotations_boxPlots} shows per-class box plots for the Intersection-over-Union (IoU) measure for all pairs of annotations. 
RB annotations demonstrate gained consistency for all classes.
For conventional annotations, we report an average IoU~=~0.63 and a standard deviation of~0.22,  whereas for RB annotations we report increased average IoU~=~0.83 with a lower standard deviation of~0.11. 

To assess how consistency changes as more annotations are collected, we employ annotators via the Amazon Mechanical Turk (MTurk) for two object interactions from BEOID, namely the actions of `scan card' and `wash cup', for which we gathered 45 conventional and RB labels.
Box plots for MTurk labels are included in Figure~\ref{fig:beoidAnnotations_boxPlots}, showing marginal improvements with RB annotations as well.
We will revisit RB annotations in detail in Section~\ref{sec:rubiconBoundaries}.

In the next Section, we assess the robustness of current action recognition approaches to variations in temporal boundaries.

\vspace*{-6pt}
\section{Temporal Bounds: Assessing Robustness}
\label{sec:experiments}
\vspace*{-4pt}

To assess the effect of temporal bounds on action recognition, we systematically vary the \textit{start} and \textit{end} times of annotated segments for the {three} datasets, and report comprehensive results on the effect of such alterations.

Results are evaluated using 5-fold cross validation. 
For training, only ground truth segments are considered.
We then classify \textit{both} the original ground truth and the generated segments. We provide results using Improved Dense Trajectories \cite{Wang2013} encoded with Fisher Vector \cite{Sanchez2013} (IDT FV)\footnote{IDT features have been extracted using GNU Parallel \cite{Tange2011a}.} and Convolutional Two-Stream Network Fusion for Video Action Recognition (2SCNN) \cite{feichtenhofer2016convolutional}. 
The encoded IDT FV features are classified with a linear SVM.
Experiments on 2SCNN are carried out using the provided code and the proposed VGG-16 architecture pre-trained on ImageNet and tuned on UCF101~\cite{soomro2012ucf101}.
We fine-tune the spatial, temporal and fusion networks on each fold's training set.

Theoretically, the two action recognition approaches are likely to respond differently to variations in start and end times.
Specifically, 2SCNN averages the classification responses of the fusion network obtained on $n$ frames randomly extracted from a test video $v$ of length $|v|$. In our experiments, $n = \min(20, |v|)$. Such strategy should ascribe some degree of resilience to 2SCNN. 
IDT \Davide{densely} samples feature points in the first frame of the video, whereas in the following frames only new feature points are sampled to replace the missing ones. This entails that IDT FV should be more sensitive to start (specifically) and end time variations, at least for shorter videos.
This fundamental difference makes both approaches interesting to assess for robustness.

\begin{table}[t]
	\centering
	\resizebox{1\columnwidth}{!}{
		\begin{tabular}{l|c|c|c}
			\hline
			Dataset 	& N. of $gt$ segments & N. of $gen$ segments & Classes \\ \hline
			BEOID~\cite{Damen2014a} & 742				  & 16691				 & 34 \\
			GTEA Gaze+~\cite{Fathi2012} & 	1141		  & 22221				 & 42 \\ 
			CMU~\cite{de2008guide} & 	450		     	  & 26160   			 & 31 \\ \hline
	\end{tabular}}
	\caption{Number of ground truth/generated segments and number of classes for BEOID, GTEA Gaze+ and CMU.}
	\label{table:datasetsInfo}
\end{table}

\begin{figure}[t]
	\centering
	\includegraphics[width=0.8\columnwidth]{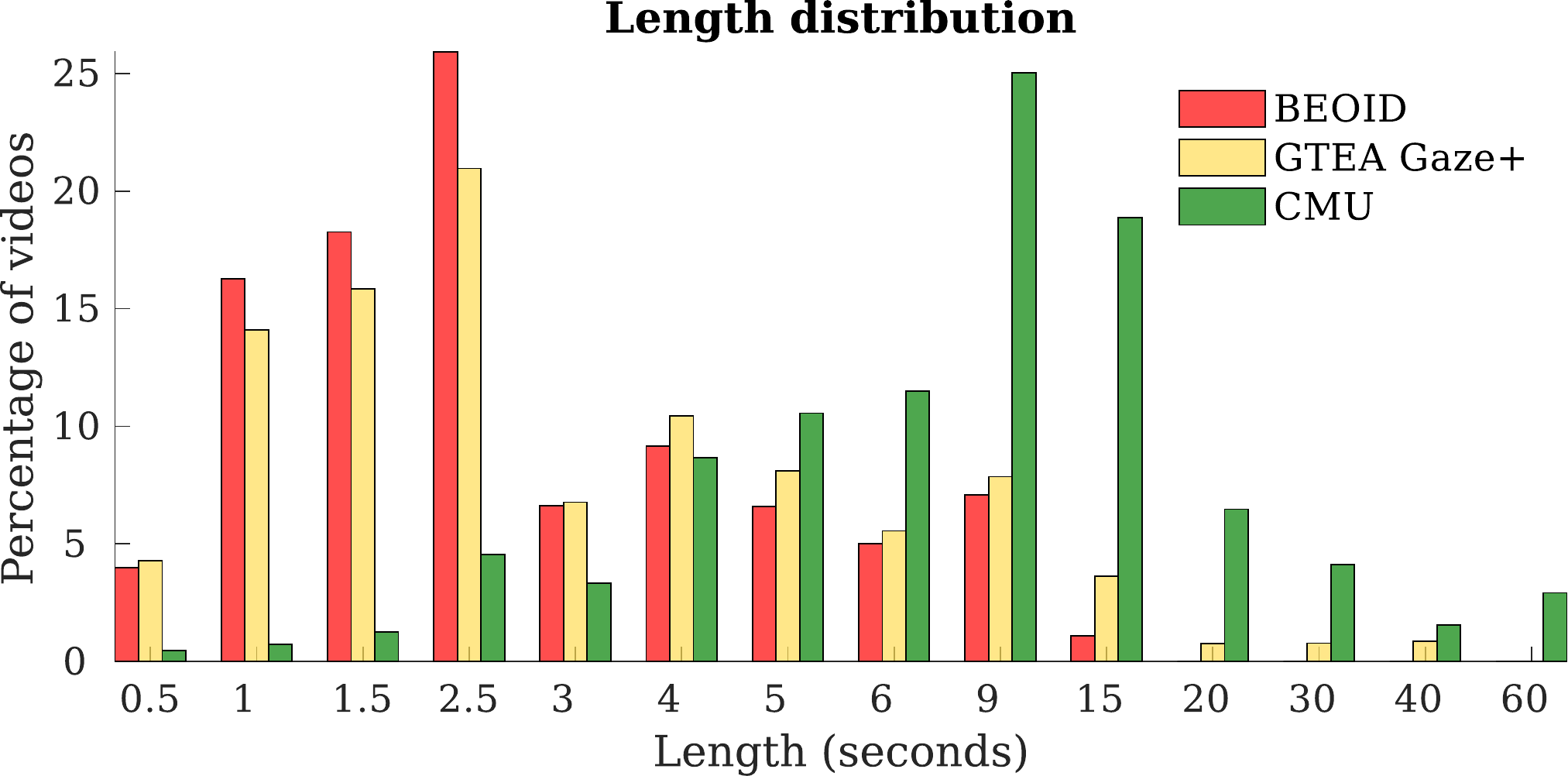}
	\label{fig:datasetLengthDistribution}
	\caption{Video's length distribution {across datasets}.}
	\label{fig:datasetLengthDistribution}
\end{figure}

\begin{figure*}[t]
	\centering
	\includegraphics[width=\textwidth]{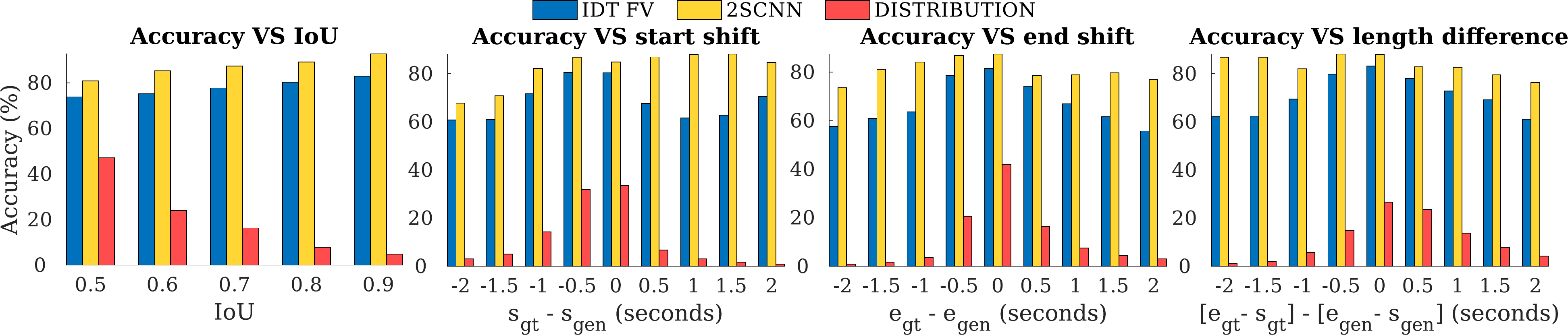}
	\caption{BEOID: classification accuracy vs IoU, start/end shifts and length difference between $gt$ and generated segments.}
	\label{fig:beoidResults}
\end{figure*}

\begin{figure*}[t!]
	\centering
	\includegraphics[width=\textwidth]{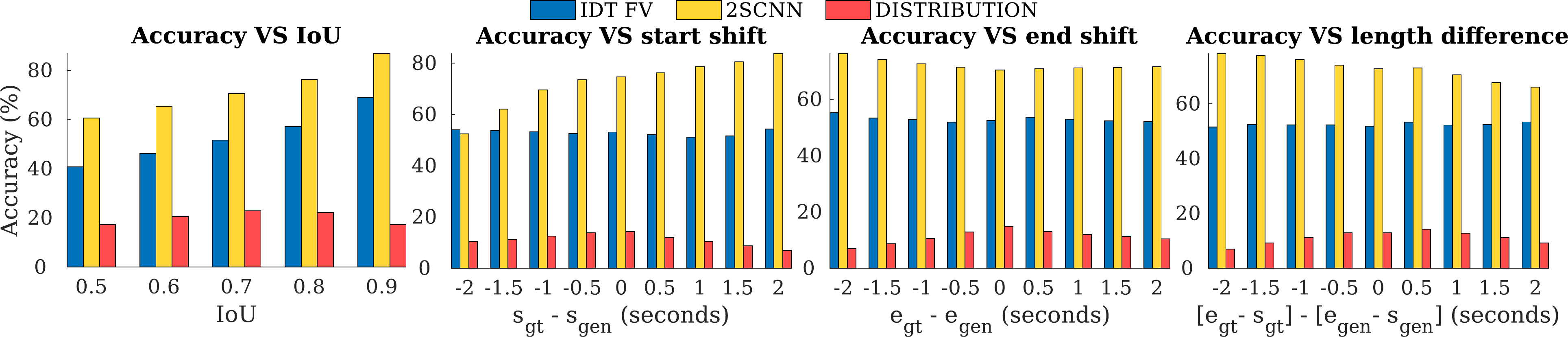}
	\caption{CMU: classification accuracy vs IoU, start/end shifts and length difference.}
	\label{fig:cmuResults}
\end{figure*}

\subsection{Generating Segments}
\label{subsec:generatingSegments}

Let $v_{gt}$ be a ground truth action segment obtained by cropping an untrimmed video from time $s_{gt}$ to time $e_{gt}$, which denote the annotated ground truth start and end times. 
We vary both $s_{gt}$ and $e_{gt}$ in order to generate new action segments with different temporal bounds. More precisely, let ${s_{gen}^{0} = s_{gt} - \Delta}$ and let $s_{gen}^{n} = s_{gt} + \Delta$. The set containing candidate start times is defined as:
\begin{equation*}
\mathcal{S} = \{s_{gen}^{0}, s_{gen}^{0} + \delta, s_{gen}^{0} + 2\delta, ..., s_{gen}^{0} + (n-1) \delta, s_{gen}^{n}\}
\end{equation*}
Analogously, let $e_{gen}^{0} = e_{gt} - \Delta$ and let $e_{gen}^{n} = e_{gt}~+~\Delta$, the set containing candidate end times is defined as:
\begin{equation*}
\mathcal{E} = \{e_{gen}^{0}, e_{gen}^{0} + \delta, e_{gen}^{0} + 2\delta, ..., e_{gen}^{0} + (n-1) \delta, e_{gen}^{n}\}
\end{equation*}
To accumulate the set of generated action segments, we take all possible combinations of $s_{gen} \in \mathcal{S}$ and $e_{gen} \in \mathcal{E}$ and keep only those such that the Intersection-over-Union between $[s_{gt}, e_{gt}]$ and $[s_{gen}, e_{gen}] \geq 0.5$. In all our experiments, we set $\Delta = 2$ and $\delta = 0.5$ seconds.

\subsection{Comparative Evaluation}
\label{subsec:results}

Table \ref{table:datasetsInfo} reports the number of ground truth and generated segments for BEOID, GTEA Gaze+ and CMU. Figure \ref{fig:datasetLengthDistribution} illustrates the segments' length distribution for the three datasets, showing considerable differences: BEOID and GTEA Gaze+ contain mostly short segments (1-2.5 seconds), although the latter includes also videos up to 40 seconds long. CMU has longer segments, with the majority ranging from {5 to 15 seconds.} %We next report, for each dataset, accuracy vs overlap (IoU), start/end shifts and length difference between the ground truth and the generated segments.

\begin{figure*}[t]
	\centering
	\includegraphics[width=\textwidth]{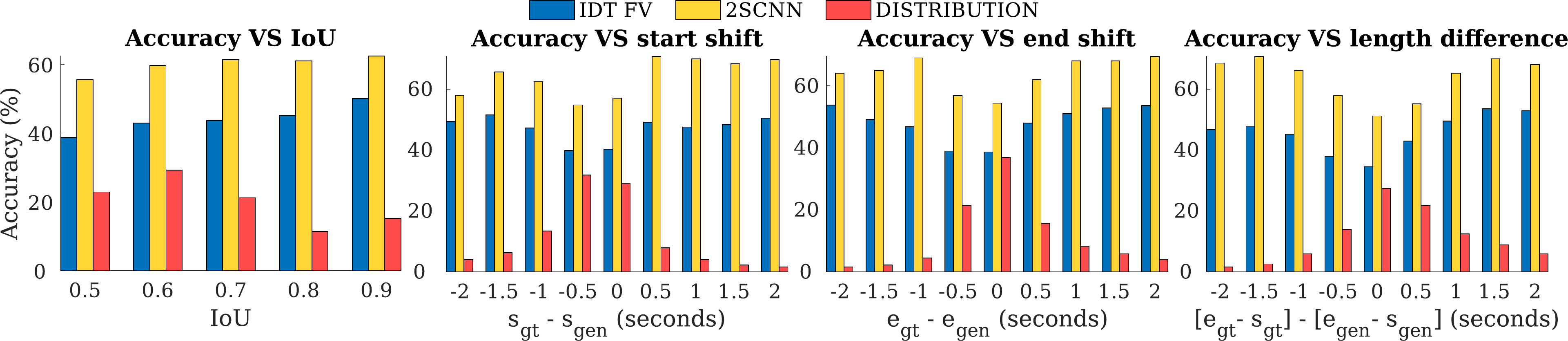}
	\caption{GTEA Gaze+: classification accuracy vs IoU, start/end shifts and length difference.} \label{fig:gteaResults}
\end{figure*}

\begin{figure*}[h!]
	\centering
	\includegraphics[width=0.9\textwidth]{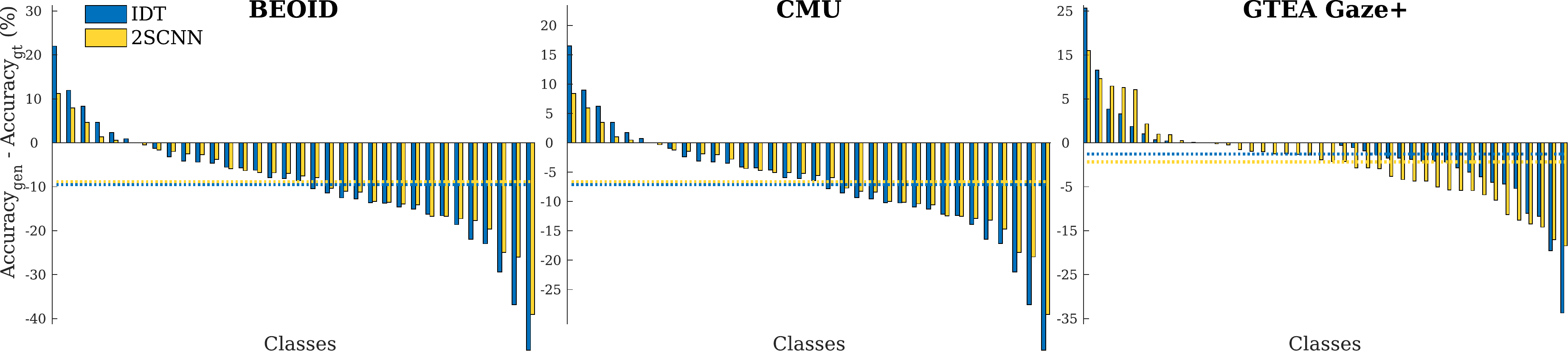}
	\caption{Accuracy per class differences. Most classes exhibit a drop in accuracy when testing generated segments.}
	\label{fig:classAccuracyComparison}
\end{figure*}

\noindent{\textbf{BEOID~\cite{Damen2014a}} \hspace{4pt}} is the evaluated dataset with the most consistent and the tightest temporal bounds. When testing the ground truth segments using both IDT FV and 2SCNN, we observe high accuracy for ground truth segments - respectively 85.3\% and 93.5\% - as shown in Table~\ref{table:accuracyResultsAll}. When classifying the generated segments, we observe a drop in accuracy of 9.9\% and 9.7\% respectively.

Figure \ref{fig:beoidResults} shows detailed results where accuracy is reported vs IoU, start/end shifts and length difference between ground truth and generated segments.
We particularly show the results for shifts in the start and the end times independently. 
A \textit{negative start shift} implies that a generated segment begins before the corresponding ground truth segment, and a \textit{negative end shift} implies that a generated segment finishes before the corresponding ground truth segment.
These terms are used consistently throughout this section.
Results show that: (i) as IoU decreases the accuracy drops consistently for IDT FV and 2SCNN - which questions both approaches' robustness to temporal bounds alterations; (ii) IDT FV exhibits lower accuracy with both negative and positive {start/end} shifts; (iii)~IDT FV similarly exhibits lower accuracy with negative and positive length differences. This is justified as BEOID segments are {tight; by} expanding a segment we include new potentially noisy or irrelevant frames that confuse the classifiers;
%, particularly as IDT sampling strategy relies on the segment's first frame
(iv) 2SCNN is more robust to length difference which is understandable as it randomly samples a maximum of 20 frames regardless of the length. 
While these are somehow expected, we also note that (v)~2SCNN is robust to \textit{positive start shifts}.

\begin{table}[t]
\centering
\resizebox{1\columnwidth}{!}{
\begin{tabular}{l|c|c||c|c}
\hline
Dataset &  $\textcolor{idtColor}{\text{IDT FV}}_{gt}$  & $\textcolor{idtColor}{\text{IDT FV}}_{gen}$ & $\textcolor{cnnColor}{\text{2SCNN}}_{gt}$ & $\textcolor{cnnColor}{\text{2SCNN}}_{gen}$ \\ \hline
BEOID & 85.3         & 75.4         & 93.5       & 83.8        \\ 
CMU   & 54.9         & 52.8         & 76.0       & 71.7        \\ 
GTEA Gaze+ & 45.4         & 43.3         & 61.2       & 59.6   \\ \hline
\end{tabular}}
\caption{Classification accuracy for ground truth and generated segments for BEOID, CMU and GTEA~Gaze+.}
\label{table:accuracyResultsAll}
\end{table}

% \begin{table}[t]
% 	\centering
% 	\resizebox{1\columnwidth}{!}{
% 		\begin{tabular}{l|C{1.45cm}|C{1.45cm}||C{1.45cm}|C{1.45cm}C{0cm}C{1.45cm}|C{1.45cm}}
% 			\cline{1-5} \cline{7-8}
% 			Dataset & $\textcolor{idtColor}{\text{IDT FV}}_{gt}$  & $\textcolor{idtColor}{\text{IDT FV}}_{gen}$ & $\textcolor{cnnColor}{\text{2SCNN}}_{gt}$ & $\textcolor{cnnColor}{\text{2SCNN}}_{gen}$ & & $\textcolor{cnnColor}{\text{2SCNN}}_{gt}^{aug}$ & $\textcolor{cnnColor}{\text{2SCNN}}_{gen}^{aug}$\\ \cline{1-5} \cline{7-8}
% 			BEOID & 85.3         & 75.4         &93.5       & 83.8  & & 92.3 & 86.6 \\ 
% 			CMU   & 54.9         & 52.8         & 76.0       & 71.7  & & - & -     \\ 
% 			GTEA Gaze+ & 45.4         & 43.3         & 61.2       & 59.6 & & 57.9 & 58.1   \\ \cline{1-5} \cline{7-8}
			
% 	\end{tabular}}
% 	\caption{Classification accuracy for ground truth and generated segments for BEOID, CMU and GTEA~Gaze+. N{[No, unfortunately this is not working. You need to remove the data augmentation from here as it's confusing, table too small and you don't have it for all datasets... Data augmentation should be added as a small table under the new section on data augmentation. We can worry about space later but at this moment this kills the main table of results]}}
% 	\label{table:accuracyResultsAll}
% \end{table}

\noindent{\textbf{CMU~\cite{de2008guide}} \hspace{4pt}} is the dataset with longer ground truth segments.
Table \ref{table:accuracyResultsAll} compares results obtained for CMU's ground truth and generated segments. 
For this dataset, IDT FV accuracy drops by 2.1\% for the generated segments, whereas 2SCNN drops by 4.3\%. 
In Figure \ref{fig:cmuResults}, CMU consistently shows low robustness for both IDT FV and 2SCNN. % to lower overlap between the ground truth and generated segments.
As IoU changes from $> 0.9$ to $> 0.5$, we observe a drop of more than 20\% in accuracy for both.
However, due to the long average length of segments in CMU, the effect of shifts in start end times as well as length differences is not visible for IDT FV.
Interestingly for 2SCNN, the accuracy slightly improves with \textit{positive start shift}, \textit{negative end shift} and \textit{negative length difference}. This suggests that CMU's ground truth bounds are somewhat loose and that tighter segments are likely to contain more discriminative frames.

\noindent{\textbf{GTEA Gaze+~\cite{Fathi2012}} \hspace{4pt}} is the dataset with the most inconsistent bounds, based on our observations. 
Table~\ref{table:accuracyResultsAll} shows that accuracy for IDT FV drops by 2.1\%, while overall accuracy for 2SCNN drops marginally (1.6\%).
This should not be mistaken for robustness, and that is evident when studying the results in Figure~\ref{fig:gteaResults}.
For all variations (i.e. start/end shifts and length differences), the generated segments achieve higher accuracy for both IDT FV and 2SCNN.
When labels are inconsistent, shifting temporal bounds does not systematically alter the visual representation of the tested segments.
%One thus concludes that t
{T}he generated segments tend to include (or exclude) frames that increase the similarity between the testing and training segments.

Figure \ref{fig:classAccuracyComparison} reports per-class differences between generated and ground truth segments. Positive values entail that the accuracy for the given class is higher when testing the generated segments, and vice versa. Horizontal lines indicate the average accuracy difference.
In total, 63\% of classes in all three datasets exhibit a drop in accuracy drop when using IDT FV compared to 80\% when using 2SCNN.

\begin{table}[h]
\centering
\resizebox{1.0\columnwidth}{!}{
\begin{tabular}{l|c|c||c|c}
\hline
Dataset & $\textcolor{cnnColor}{\text{2SCNN}}_{gt}$ & $\textcolor{cnnColor}{\text{2SCNN}}_{gen}$ & $\textcolor{cnnColor}{\text{2SCNN}}_{gt}^{\textcolor{augColor}{aug}}$ & $\textcolor{cnnColor}{\text{2SCNN}}_{gen}^{\textcolor{augColor}{aug}}$\\ \hline
BEOID & \textbf{93.5}       & 83.8   &  92.3      &   86.6     \\ 
GTEA Gaze+ & \textbf{61.2}       & 59.6  &  57.9      &   58.1 \\ \hline
\end{tabular}}
% \begin{tabular}{l|c|c}
% \hline
% Dataset    & $\textcolor{cnnColor}{\text{2SCNN}}_{gt}^{aug}$ & $\textcolor{cnnColor}{\text{2SCNN}}_{gen}^{aug}$ \\ \hline
% BEOID      &  92.3      &   86.6      \\
% GTEA Gaze+ &  57.9      &   58.1    \\  \hline
% \end{tabular}}
\caption{\Dima{2SCNN data augmentation results.}}
\label{table:dataAugmentation}
\end{table}

\noindent{\textbf{Data augmentation:} \hspace{4pt}} {For completeness{, we evaluate the performance }when using temporal data augmentation methods {on two datasets}. {Generated segments in Section~\ref{subsec:generatingSegments} are used to augment training.}
    %Augmentation was achieved by varying the start and end times of training segments, as explained in Section \ref{subsec:generatingSegments}. 
    We {double} the size of the training sets, taking random samples for augmentation. Test sets remained unvaried.
	%Training time significantly increased (2x) on both datasets. 
    Results are reported in Table \ref{table:dataAugmentation}. While we observe an increase in robustness, we also notice a drop in accuracy for ground truth segments, respectively of 1\% and 4\% for BEOID and GTEA Gaze+.%These additional results highlight that consistent labeling cannot be substituted with data augmentation, as we will discuss in the following Section.
}

In conclusion, we note that both IDT FV and 2SCNN are sensitive to changes in temporal bounds for both consistent and inconsistent annotations.
Approaches that improve robustness using data augmentation could be attempted, however a broader look at how the methods could be inherently more robust is needed, particularly for CNN architectures.

\vspace*{-6pt}
\section{{\hspace*{-1pt}Labeling} Proposal: The Rubicon Boundaries}
\label{sec:rubiconBoundaries}
\vspace*{-4pt}

The problem of defining consistent temporal bounds of an action is most akin to the problem of defining consistent bounding boxes of an object. 
Attempts to define guidelines for annotating objects' bounding boxes started nearly a decade ago.
Among others, the VOC Challenge 2007~\cite{everingham2010pascal} proposed what has become the standard for defining the bounding box of an object in images.
These consistent labels have been used to train state-of-the-art object detection and classification methods.
With this same spirit, in this Section we propose an approach to consistently segment the temporal scope of an object interaction. 

\noindent \textbf{Defining RB:} The Rubicon Model of Action Phases \cite{gollwitzer1990action}, developed in the field of Psychology, posits an action as a goal a subject desires to achieve and identifies the main sub-phases the person gets through in order to complete the action.
First, a person decides what goal he wants to obtain. After forming his intention, he enters the so-called \textit{pre-actional} phase, that is a phase where he plans to perform the action. 
Following this stage, the subject acts towards goal achievement in the \textit{actional phase}. 
The two phases are delimited by three transition points: the initiation of prior motion, the start of the action and the goal achievement. 

The model is named after the historical fact of Caesar crossing the Rubicon river, which became a metaphor for deliberately proceeding past a point of no return, which in our case is the transition point that signals the beginning of an action. 
We take inspiration from this model, specifically from the aforementioned transitions points, and define two phases for an object interaction:

\begin{figure}[!t]
	\centering
	\includegraphics[width=1.0\columnwidth]{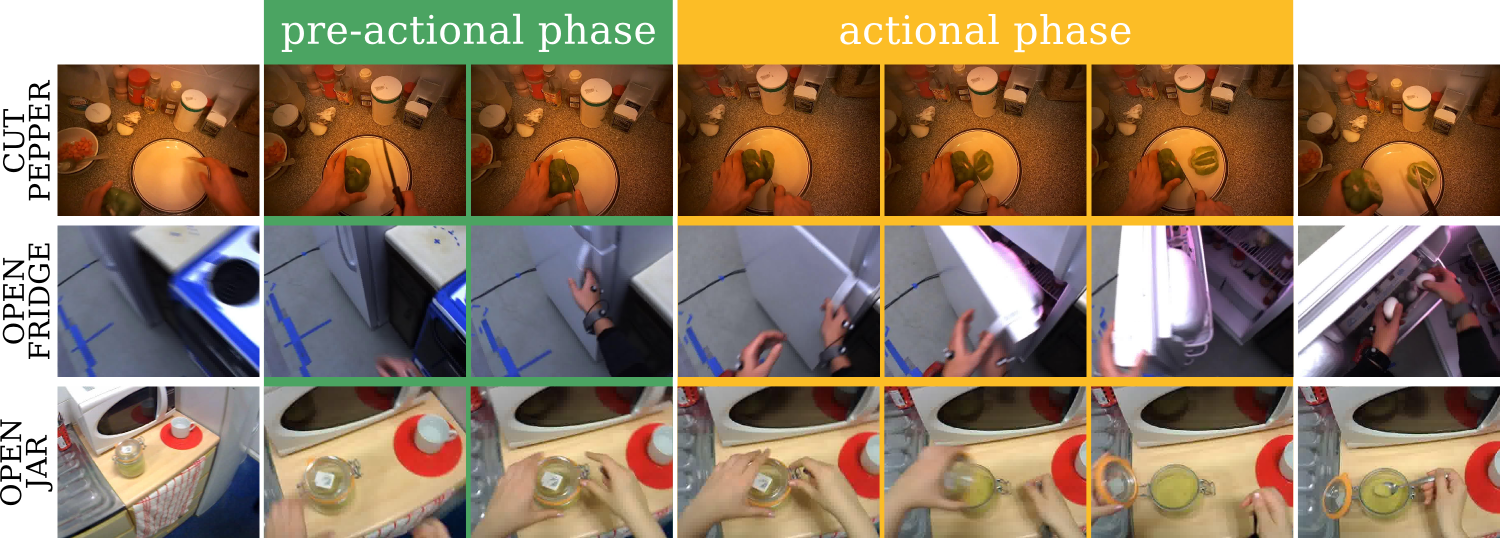}
	\caption{Rubicon Boundaries labeling examples for three object interactions.}
	\label{fig:rubiconBoundariesExample}
\end{figure}

\vspace*{2pt}
\noindent\textbf{\textit{Pre-actional phase}} This sub-segment contains the preliminary motion that directly precedes the goal, and is required for its completion. When multiple motions can be identified, the pre-actional phase should contain only the last one;\\
\textbf{\textit{Actional phase}} This is the main sub-segment containing the motion strictly related to the fulfillment of the goal. The actional phase starts immediately after the pre-actional phase.

In the following section, we refer to a label as an RB annotation when the \textit{beginning} of an object interaction aligns with the \textit{start} of the pre-actional phase and the \textit{ending} of the interactions aligns with the \textit{end} of the actional phase.

Figure \ref{fig:rubiconBoundariesExample} depicts three object interactions labeled according to the Rubicon Boundaries. The top sequence illustrates the action of cutting a pepper. 
The sequence shows the subject fetching the knife before cutting the pepper and taking it off the plate.
Based on the aforementioned definitions, the pre-actional phase is limited to the motion of moving the knife towards the pepper in order to slice it.
This is directly followed by the actional phase where the user cuts the pepper.
The actional phase ends as the goal of `cutting' is completed.
The middle sequence illustrates the action of opening a fridge, showing a person approaching the fridge, reaching towards the handle before pulling the fridge door open.
In this case, the pre-actional phase would be the reaching motion, while the actional phase would be the pulling motion. 

\begin{figure}[!t]
	\centering
	\includegraphics[width=1.0\columnwidth]{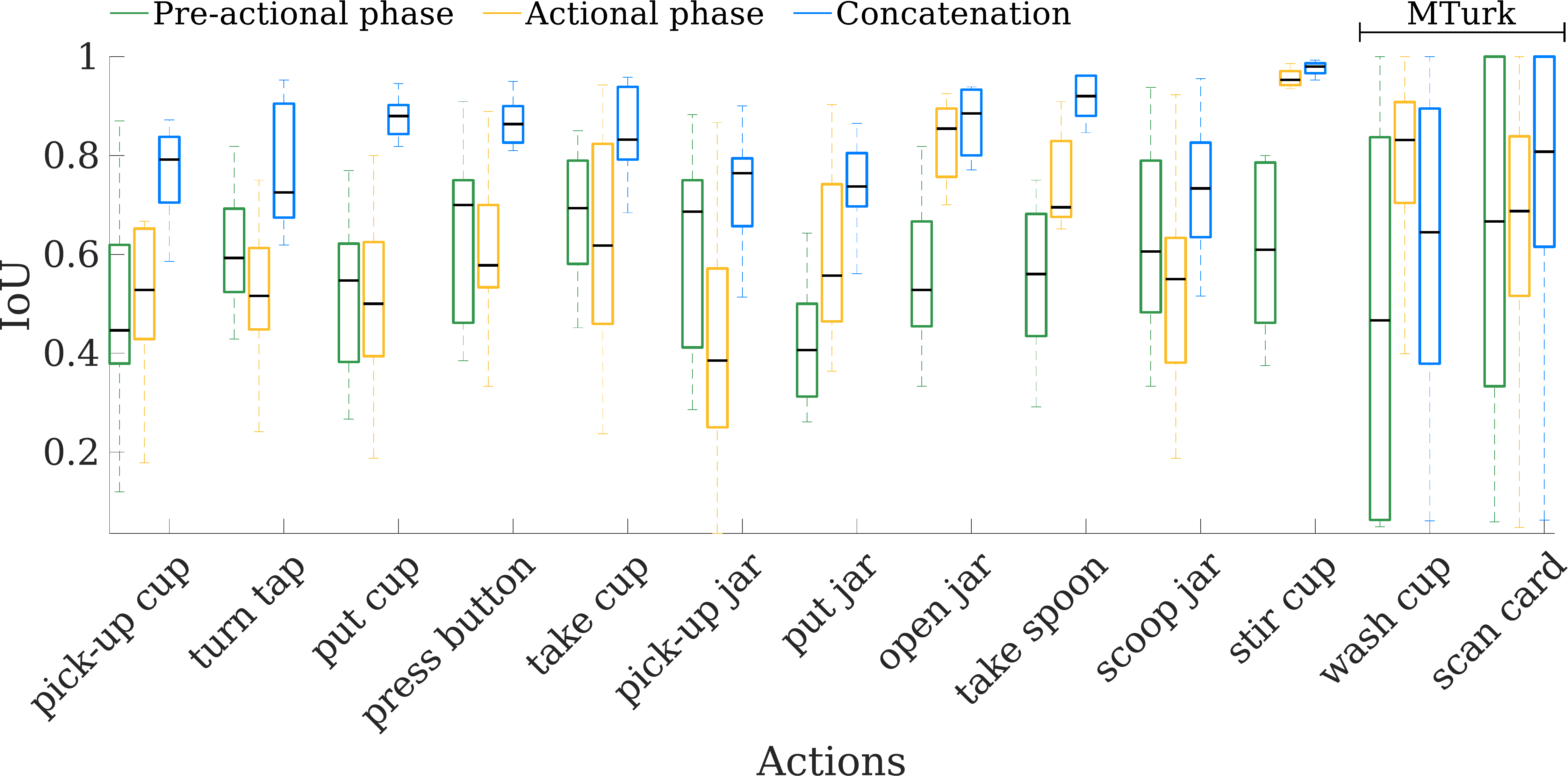}
	\vspace*{-6pt}
	\caption{IoU comparison among the pre-actional phase (green), the actional phase (yellow) and their concatenation (blue) for several object interactions of BEOID.}
	\label{fig:preActIoU}
\end{figure}

\noindent \textbf{Evaluating RB:} We evaluate our RB proposal for consistency, intuitiveness as well as accuracy and robustness. 

\noindent \textbf{(i) Consistency:} We already reported consistency results in Section \ref{subsec:multiple}, where RB annotations exhibit higher average overlap and less variation for all the evaluated object interactions - average IoU for all pairs of annotators increased from 0.63 for conventional boundaries to 0.83 for RB. 
Figure \ref{fig:preActIoU} illustrates per-class IoU box plots for the pre-actional and the actional phases separately, along with the concatenation of the two. 
For 7 out of the 13 actions, the actional phase was more consistent than the pre-actional phase, and for 12 out of the 13 actions, the concatenation of the phases proved the highest consistency.

\noindent \textbf{(ii) Intuitiveness:} While RB showed higher consistency in labeling, any new approach for temporal boundaries would require a shift in practice. We collect RB annotations from university students as well as from MTurk annotators. Locally, students successfully used the RB definitions to annotate videos with no assistance. However, this has not been the case for MTurk annotators for the two object interactions `wash cup' and `scan card'. The MTurk HIT provided the formal definition of the \textit{pre-actional} and \textit{actional} phases, then ran two multiple-choice control questions to assess the ability of annotators to distinguish these phases from a video. The annotators had to select from textual descriptions what the pre and the actional phases entailed. 
For both object interactions, only a fourth of the annotators answered the control questions correctly.

Three possible explanations could be given, namely: annotators were accustomed to the conventional labeling method {and} did not spend sufficient time to study the definitions, or the definitions were difficult to understand. Further experimentation is needed to understand the cause.

\begin{figure}[!t]
	\centering
	\includegraphics[width=1.0\columnwidth]{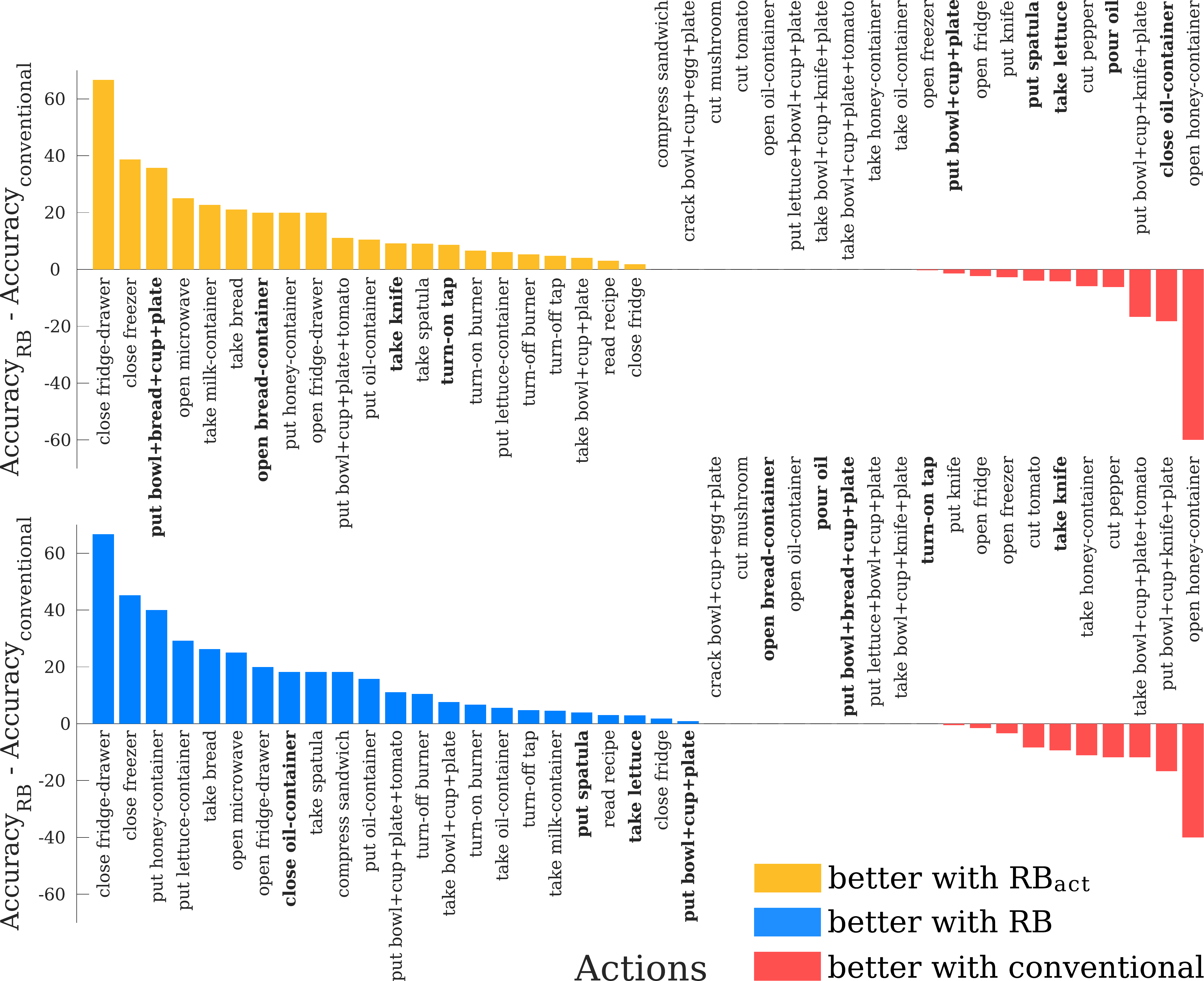}
	\caption{GTEA Gaze+: class accuracy difference between conventional and RB annotations. \Davide{Some classes achieved higher accuracy only with RB\textsubscript{act}, while other did only with the full RB segment. Bold highlights such cases.}
    %Accuracy for classes in bold changed when using either the actional phase alone or the full RB segment.
    }
	\label{fig:gteaRBClassAccuracyComparison}
\end{figure}

\noindent \textbf{(iii) Accuracy:} We annotated GTEA Gaze+ using the Rubicon Boundaries, by employing three people to label its 1141 segments\footnote{{RB labels and {video of results are available on project webpage:} \url{http://www.cs.bris.ac.uk/~damen/Trespass/}}}.
For these experiments, we asked annotators to label both the pre-actional and the actional phase. 

{In Table \ref{table:gteaRBResults2}, we} report results for the actional phase alone (RB\textsubscript{act}) as well as the concatenation of the two phases (RB){,} using 2SCNN on the same 5 folds from Section~\ref{subsec:results}.
% compares results obtained with conventional and RB labels. 
 The concatenated RB segments proved the most accurate, leading to an increase of more than 4\% in accuracy compared to conventional ground truth segments. 
{Temporal augmentation {on conventional labels ($\textcolor{originalColor}{\text{Conv}}^{\textcolor{augColor}{aug}}_{gt}$)} results in a drop of accuracy by 7.7\% compared with the RB segments, highlighting that consistent labeling cannot be substituted with data augmentation.
}
Figure \ref{fig:gteaRBClassAccuracyComparison} shows the accuracy per class difference between the two sets of RB annotations and the conventional labels. When using RB\textsubscript{act}, 21/42 classes improved, whereas accuracy dropped for 11 classes compared to the conventional annotations. When using the full RB segment, 23/42 classes improved, while 10 classes were better recognized with the conventional annotations. 
In each case, 10 and 9 classes remain unchanged.

Given that the experimental setup was identical to that used for the conventional annotations, the boost in accuracy can be ascribed solely to the new action boundaries. Indeed, the RB approach helped the annotators to more consistently segment the object interactions contained in GTEA Gaze+, which is one of the most challenging datasets for egocentric action recognition. 

\noindent \textbf{(iv) Robustness:} 
Table \ref{table:gteaRBResults2} {also} compares the newly annotated RB segments to generated segments with varied start and end times, as explained in Section~\ref{subsec:generatingSegments}.
While RB$_{gen}$ shows higher accuracy than the Conventional$_{gen}$ segments (59.6\% as reported in Table~\ref{table:accuracyResultsAll}), we still observe a clear drop in accuracy between $gt$ and $gen$ segments. Interestingly, we {observe} improved robustness when using the actional phase alone. Given that the actional segment's start is closer in time to the beginning of the object interaction, when varying the start of the segment we are effectively including part of the pre-actional phase in the generated segment, which assists in making actions more discriminative.%, as shown in Table~\ref{table:gteaRBResults2}.

Importantly, we {show} that RB annotations improved both consistency and accuracy of annotations on the largest dataset of egocentric object interactions. \Dima{We believe these} form solid basis for further discussions and experimentation on consistent labeling of temporal boundaries.

\vspace*{-6pt}
\section{Conclusion and Future Directions}
\label{sec:conclusion}
\vspace*{-4pt}
% \begin{table}[t]
% \centering
% \resizebox{1\columnwidth}{!}{
% \begin{tabular}{C{2.8cm}|C{2.8cm}|C{2.8cm}}
% \hline
% \textcolor{originalColor}{Conventional}& \textcolor{cnnColor}{RB\textsubscript{act}}& \textcolor{rbColor}{RB}\\ \hline
% 61.2 & 64.9 & \textbf{65.6}  \\ \hline
% \end{tabular}}
% \caption{GTEA Gaze+: 2SCNN classification accuracy comparison for ground truth segments between conventional and RB annotations.}
% \label{table:gteaRBResults}
% \end{table}

% \begin{table}[t]
% \centering
% \resizebox{1\columnwidth}{!}{
% \begin{tabular}{C{2cm}|C{2cm}||C{2cm}|C{2cm}}
% \hline
% \textcolor{cnnColor}{RB\textsubscript{act}} $gt$ & \textcolor{cnnColor}{RB\textsubscript{act}} $gen$ & \textcolor{rbColor}{RB} $gt$ & \textcolor{rbColor}{RB} $gen$ \\ \hline
% 64.9 & 63.2 & 65.6  & 61.7  \\ \hline
% \end{tabular}}
% \caption{GTEA Gaze+: 2SCNN classification accuracy comparison for ground truth ($gt$) and generated ($gen$) segments obtained with RB annotations.}
% \label{table:gteaRBResults2}
% \end{table}

\begin{table}[t]
	\centering
	\resizebox{1\columnwidth}{!}{
		\begin{tabular}{C{1.2cm}|C{1.2cm}||C{1.2cm}|C{1.2cm}||C{1.2cm}|C{1.2cm}}
			\hline
			$\textcolor{originalColor}{\text{Conv}}_{gt}$ & $\textcolor{originalColor}{\text{Conv}}^{\textcolor{augColor}{aug}}_{gt}$ & $\textcolor{cnnColor}{\stackrel{\text{act}}{\text{RB}}}_{gt}$ & $\textcolor{cnnColor}{\stackrel{\text{act}}{\text{RB}}}_{gen}$ & $\textcolor{rbColor}{\text{RB}}_{gt}$ & $\textcolor{rbColor}{\text{RB}}_{gen}$ \\ \hline
			61.2 & 57.9 & 64.9 & 63.2 & \textbf{65.6}  & 61.7  \\ \hline
	\end{tabular}}
	\caption{{GTEA Gaze+: 2SCNN classification accuracy comparison for conventional annotations (ground truth and augmented) and RB labels (ground truth and generated).}}
	\label{table:gteaRBResults2}
\end{table}

Annotating temporal bounds for object interactions is the base for supervised action recognition algorithms.
In this work, we uncovered inconsistencies in temporal bound annotations within and across three egocentric datasets. We assessed the robustness of both hand-crafted features and fine-tuned end-to-end recognition methods, and demonstrated that both IDT FV and 2SCNN are susceptible to variations in start and end times. We then proposed an approach to consistently label temporal bounds for object interactions. We foresee three potential future directions:

\noindent{\textbf{Other NN architectures?} \hspace{2pt}} 
%The results demonstrate that 
\Dima{W}hile 2SCNN randomly samples frames from a video segment, the classification accuracy is still sensitive to variations in temporal bounds.
Other architectures, particularly those that model temporal progression using Recurrent networks (including LSTM), rely on labeled training samples and would thus equally benefit from consistent labeling.
Evaluating the robustness of such networks is an interesting future direction.

\noindent{\textbf{How can robustness to temporal boundaries be achieved?} \hspace{2pt} Classification methods that are inherently robust to temporal boundaries, while learning from supervised annotations, is a topic for future directions. As deep architectures reportedly outperform hand-crafted features and other classifiers, architectures that are designed to handle variations in start and end times are desired.}

\noindent{\textbf{Which temporal granularity?} \hspace{2pt}} \Dima{T}he Rubicon Boundaries address consistent labeling of temporal bounds, \Dima{but} they do not address the concern of granularity of the action. Is the action of cutting a whole tomato composed of several short cuts or is it one long action? The Rubicon Boundaries model discusses actions relative to the goal a person wishes to accomplish. The granularity of an object interaction is another matter, and annotating the level of granularity consistently has not been addressed yet. Expanding Rubicon Boundaries to enable annotating the granularity would require further investigation.

\noindent \textbf{Data Statement \& Ack:} \hspace{4pt} {Public datasets were used in this work; no new data were created as part of this study. RB annotations are available on the project's webpage. Supported by EPSRC DTP and EPSRC LOCATE (EP/N033779/1).}

{\small
\bibliographystyle{ieee}
\bibliography{references}
}

\end{document}